\documentclass[conference]{IEEEtran}
\IEEEoverridecommandlockouts

\usepackage{amsmath,amssymb,amsfonts}
\usepackage{graphicx}
\usepackage{textcomp}
\usepackage{xcolor}
\def\BibTeX{{\rm B\kern-.05em{\sc i\kern-.025em b}\kern-.08em
    T\kern-.1667em\lower.7ex\hbox{E}\kern-.125emX}}

\usepackage{graphics}           
\usepackage{times}              
\usepackage{amsmath}            
\usepackage{amssymb}            
\usepackage{graphicx}
\usepackage{tikz}
\usetikzlibrary{arrows.meta,positioning,calc}
\usepackage{algorithm}
\usepackage[noend]{algpseudocode}
\usepackage{booktabs}
\usepackage{multirow}
\usepackage{color}
\usepackage[nocompress]{cite} 
\definecolor{instructioncolor}{rgb}{.5,.5,.5}

\usepackage[font=small]{caption}

\def\eqref#1{Eq.~(\ref{#1})}

\makeatletter
\usepackage{xspace}
\DeclareRobustCommand\onedot{\futurelet\@let@token\@onedot}
\def\@onedot{\ifx\@let@token.\else.\null\fi\xspace}

\makeatother

\usepackage{array}
\newcolumntype{L}[1]{>{\raggedright\let\newline\\\arraybackslash\hspace{0pt}}m{#1}}
\newcolumntype{C}[1]{>{\centering\let\newline\\\arraybackslash\hspace{0pt}}m{#1}}
\newcolumntype{R}[1]{>{\raggedleft\let\newline\\\arraybackslash\hspace{0pt}}m{#1}}

\renewcommand{\b}[1]{\mbox{\boldmath$#1$}}

\renewcommand{\d}[1]{\b {#1}}

\usepackage{subcaption}

\title{\LARGE \bf UltraLight Luma: A Novel Edge-Deployable Perception Network for Crop-Row Segmentation in Agricultural Robotics

}

\author{Dhanushka Balasingham$^{1}$, Shamod Ginigaddarage$^{2}$, Hanojhan Rajahrajasingh$^{3}$, Nuwan Kodagoda$^{4}$ \\ and Rajitha de Silva$^{5}$%
\thanks{(Corresponding author: Dhanushka Balasingham.)}%
\thanks{$^{1}$Dhanushka Balasingham, $^{2}$Shamod Ginigaddarage, $^{3}$Hanojhan Rajahrajasingh and $^{4}$Nuwan Kodagoda are with the Faculty of Computing, Sri Lanka Institute of Information Technology, Malabe, Sri Lanka. Email: $^{1}$dhanushkasandeepa@gmail.com; $^{2}$shamodpasindu@gmail.com}%
\thanks{$^{5}$Rajitha de Silva is with Lincoln Centre for Autonomous Systems (L-CAS), University of Lincoln, UK. Email: $^{5}$odesilva@lincoln.ac.uk}%
}

\begin{document}
\maketitle
\thispagestyle{empty}
\pagestyle{empty}

\begin{abstract}
  Reliable crop-row perception is essential for autonomous agricultural robots, but low-cost deployment is constrained by computation, memory, power, and inference time. This paper addresses crop-row detection and navigation-line extraction using UltraLight Luma, a compact encoder--decoder segmentation network with a progressive learning strategy and only 13.1k trainable parameters. The predicted crop-row masks are processed using a post-processing algorithm to estimate row alignment for downstream visual servoing. UltraLight Luma improves parameter efficiency over U-Net, YOLOv8, and YOLOv26 while maintaining reliable crop-row detection performance. The model required only 25.23 mJ per inference, demonstrating its suitability for resource-constrained agricultural robots.
\end{abstract}

\noindent\textbf{Keywords: Crop-row segmentation, agricultural robotics, edge deployment, lightweight neural networks, visual servoing, Triangle Scan Method}

\section{Introduction}
\label{sec:intro}


Autonomous agricultural robots can reduce labor demand and improve field productivity, but reliable crop-row detection remains essential for navigation, localization, weeding, spraying, and inspection. Deployment on low-cost mobile platforms is constrained by computation, battery capacity, and hardware cost, while high-computation models often require expensive processors and consume more power \cite{sze2017efficient}. Therefore, crop-row detection must be accurate, efficient, cost effective, and fast enough for real-time operation.

In this work, crop-row detection is formulated as an image segmentation task using forward-facing red-green-blue (RGB) images from mobile agricultural robots. The goal is to identify visible crop-row regions for row following and navigation under practical field challenges such as shadows, weeds, missing plants, uneven soil, and perspective distortion. The target platform is a low-cost, compute-constrained embedded device with limited memory, power, and inference time.

The proposed approach uses an ultra-lightweight segmentation model to predict crop-row regions from RGB images in real time on constrained hardware. Unlike classical thresholding or line-fitting methods, it is designed to be more robust to changes in illumination, plant density, and row structure, while avoiding the high memory, power, and processor requirements of heavier deep networks \cite{sze2017efficient,enet}. Thus, the method balances accuracy, speed, and deployability for low-cost agricultural robots.

The key contributions of this paper are: (1) A novel ultra-lightweight encoder-decoder segmentation network, UltraLight Luma, is proposed with a progressive learning strategy for crop-row segmentation on low-cost agricultural robotic platforms using only 13.1k trainable parameters. (2) A navigation-line extraction module based on the Triangle Scan Method (TSM) is applied to the segmented mask to estimate the crop-row navigation line. (3) A comparative analysis against state-of-the-art models is conducted through experimental assessment.

\section{Related Work}
\label{sec:related}


Vision-based crop-row detection has used traditional image-processing methods \cite{10593228}, which are efficient but sensitive to illumination and field variability. Deep learning improves robustness, including YOLO-based bounding box methods \cite{9644466, boundingBoxYolo}, but these require centroid extraction and line fitting, which can introduce errors under canopy overlap and weed interference \cite{LI2024109264}. In contrast, semantic segmentation provides pixel-wise crop-row localization for more accurate row-geometry estimation \cite{l1,10021091}.

Multiple semantic segmentation models have been explored for crop-row detection in agricultural environments. Among them, U-Net–based approaches have demonstrated effective pixel-wise segmentation performance for identifying crop-rows under varying field conditions \cite{10021091, l1}. For example, Doha et al. reported competitive mean Intersection over Union (IoU) scores using U-Net architectures for crop-row segmentation tasks. However, conventional U-Net architectures consist of deep encoder–decoder networks with numerous convolutional layers, resulting in increased computational complexity and inference time. Consequently, more computationally efficient architectures such as LinkNet have been investigated to enable real-time agricultural navigation applications \cite{10021091}.

Although LinkNet is more computationally efficient than U-Net, it still contains approximately 11.5 million parameters \cite{linknet}. To further reduce computational cost, efficient architectures such as Efficient Neural Network (ENet) \cite{enet}, YOLO-based models \cite{agriculture15070796}, and MobileNet-based backbones \cite{s26103142} have been investigated for agricultural vision tasks, including crop-row identification. These models have demonstrated promising performance. However, their parameter counts remain relatively high for extremely constrained embedded deployment. To the best of current knowledge, existing lightweight approaches for crop-row detection have not achieved a parameter count below 0.1 million. In contrast, the proposed model uses only 13.1k parameters for crop-row segmentation, providing a substantially more compact alternative for low-cost robotic platforms.

\section{Methodology}
\label{sec:main}

\begin{figure}[h]
  \centering
  \includegraphics[width=\columnwidth]{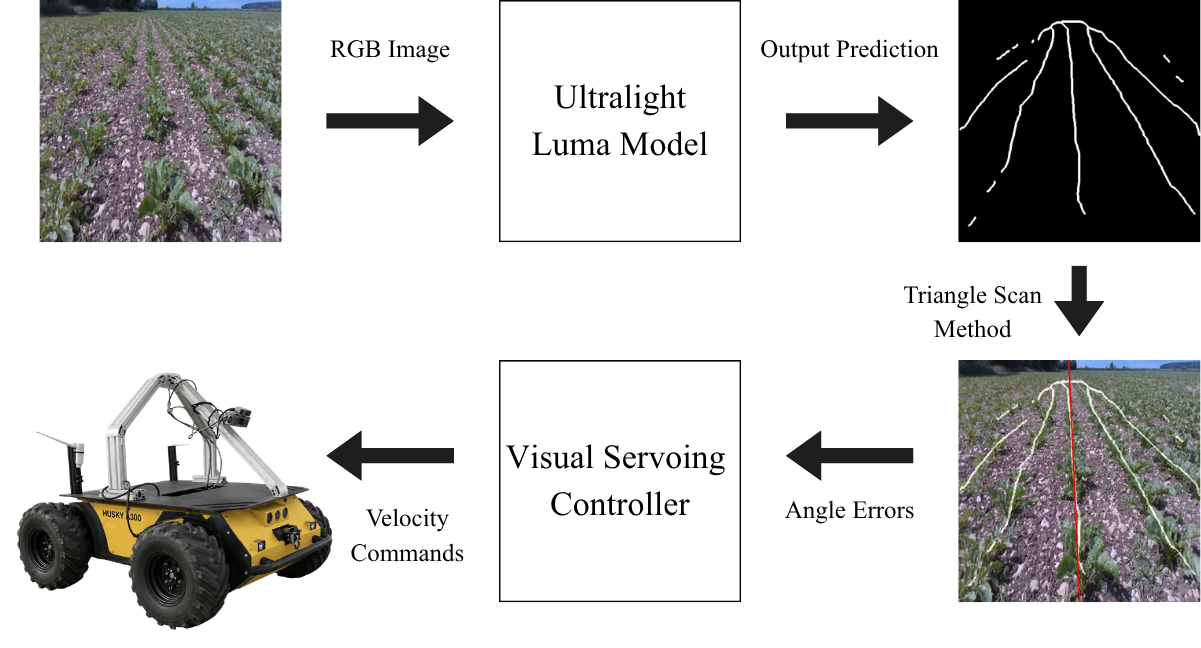}
  \caption{Overall vision-based navigation pipeline, including RGB image input, lightweight segmentation, crop-row mask prediction, and navigation-line extraction.}
  \label{fig:ultralight_luma_pipeline}
\end{figure}

Figure~\ref{fig:ultralight_luma_pipeline} presents the complete crop-row detection pipeline, where
a forward-facing camera captures RGB images, which are passed to UltraLight Luma
for crop-row segmentation. The resulting mask is processed by the TSM to estimate
the angular error, which is then fed to the visual-servoing controller to generate
velocity commands. This work focuses on two tasks: crop-row detection and navigation-line extraction for the straddling robot.

Crop-row detection is formulated as a binary semantic segmentation problem. Given an RGB image $I \in \mathbb{R}^{H \times W \times 3}$, the objective is to predict a binary mask $M \in \{0,1\}^{H \times W}$, where pixels with label 1 correspond to crop-row regions and pixels with label 0 correspond to background. The predicted crop-row mask is then used as a perception output for downstream row following and navigation. The complete crop-row perception pipeline consists of four phases.


\subsection{Dataset Preparation}
\label{sec:ds}

The dataset in \cite{l1}, collected with a front-mounted camera on a Husky robot, contains 2000 images across two crops, 11 field variations, and 50 data classes, split into 1250 training, 250 validation, and 500 test images. Each image has a binary ground-truth mask, with white pixels representing crop regions and black pixels representing background. Images and masks are resized to $64 \times 64$, and training uses spatial and photometric augmentation while preserving image--mask correspondence.

\subsection{UltraLight Luma Architecture}

\begin{figure*}[t]
  \centering
  \input{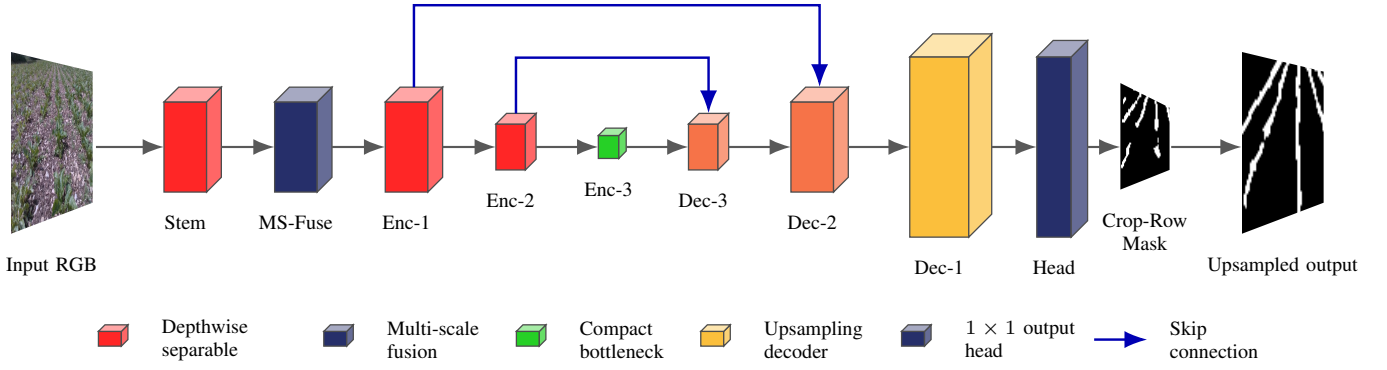}
  \caption{Architecture of the proposed UltraLight Luma encoder--decoder network for crop-row segmentation. The model uses compact depthwise-separable convolutional blocks, multi-scale feature fusion, and skip connections to preserve spatial row structure while maintaining edge-deployable complexity.}
  \label{fig:ultralight_luma_architecture}
\end{figure*}

UltraLight Luma uses a U-Net-inspired encoder--decoder structure \cite{unet} and extends the lightweight depthwise-separable design of UltraLight+ \cite{ginigaddarage2025} to segmentation. Standard convolutions are replaced with depthwise-separable (DW-Sep) blocks to reduce parameters and activation memory for edge deployment. Each DW-Sep block applies a $3\times3$ depthwise convolution followed by a $1\times1$ pointwise convolution, with Batch Normalization (BN) and ReLU after each step, as shown in Eq.~\ref{eq:dwsep}.

\begin{equation}
\label{eq:dwsep}
    \mathbf{y} = \mathrm{ReLU}\!\left(\mathrm{BN}\!\left(
    \mathrm{PW}_{1\times1}\!\left(\mathrm{ReLU}\!\left(
    \mathrm{BN}\!\left(\mathrm{DW}_{3\times3}(\mathbf{x})
    \right)\right)\right)\right)\right)
\end{equation}

where $\mathbf{x} \in \mathbb{R}^{H\times W\times C_\text{in}}$ is the input
feature map of spatial dimensions $H\times W$ with $C_\text{in}$ input channels,
$\mathrm{DW}_{3\times3}$ denotes a $3\times3$ depthwise convolution applying one
filter per input channel, $\mathrm{PW}_{1\times1}$ denotes a $1\times1$ pointwise
convolution that mixes information across channels, $\mathrm{BN}$ denotes Batch
Normalisation, $\mathrm{ReLU}$ is the rectified linear unit activation function,
and $\mathbf{y} \in \mathbb{R}^{H\times W\times C_\text{out}}$ is the output
feature map with $C_\text{out}$ output channels. This composite block is referred
to as $\mathrm{DWSep}(\cdot)$ and the full layer sequence is summarised in Table \ref{tab:arch}.

\begin{table}[h]
  \renewcommand{\arraystretch}{1.3}
  \caption{UltraLight Luma layer summary.}
  \label{tab:arch}
  \centering
  \setlength{\tabcolsep}{6pt}
  \begin{tabular}{|l|l|c|c|}
    \hline
    \textbf{Stage} & \textbf{Name} & \textbf{Output Shape} &
    \textbf{Channels} \\
    \hline
    \multirow{4}{*}{Encoder}
      & Input   & $64\!\times\!64$ &  3 \\
      & Stem    & $32\!\times\!32$ & 16 \\
      & MS-Fuse & $32\!\times\!32$ & 32 \\
      & Enc-1   & $32\!\times\!32$ & 32 \\
    \hline
    \multirow{2}{*}{Downsample}
      & Enc-2   & $16\!\times\!16$ & 32 \\
      & Enc-3   &  $8\!\times\!8$  & 32 \\
    \hline
    \multirow{3}{*}{Decoder}
      & Dec-3   & $16\!\times\!16$ & 32 \\
      & Dec-2   & $32\!\times\!32$ & 16 \\
      & Dec-1   & $64\!\times\!64$ & 16 \\
    \hline
    Output
      & Seg-Out & $64\!\times\!64$ &  2 \\
    \hline
  \end{tabular}
\end{table}

The network processes inputs through four sequential stages namely encoder, downsample, decoder, and output head.

\subsubsection{Encoder}

The encoder stage begins with a stem convolution that applies a standard
$3\times3$ convolution with stride 2, reducing the $64\times64\times3$ input
to a $32\times32\times16$ feature map while extracting initial low-level
spatial features. The stem output $\mathbf{x}_\text{stem}$ is then passed
to a multi-scale fusion block that processes features at two spatial scales
in parallel. The first branch preserves the current resolution using a
$3\times3$ depthwise convolution, while the second branch applies a strided
depthwise convolution and upsamples the result back to $32\times32$. The
two branches are concatenated and fused through a $1\times1$ pointwise
convolution. The fused representation is then processed by a DW-Sep block
(Enc-1), whose output is stored as skip connection that is denoted by
$\mathbf{s}_1 \in \mathbb{R}^{32\times32\times32}$ for later use in the
decoder. The encoder stage is expressed in \ref{eq:stage1}.

\begin{align}
\mathbf{b}_1 &= \mathrm{DW}_{3\times3}(\mathbf{x}_\text{stem}), \\
\mathbf{b}_2 &= \mathrm{Up}\!\left(\mathrm{DW}_{3\times3,s=2}
(\mathbf{x}_\text{stem})\right), \\
\mathbf{s}_1 &= \mathrm{DWSep}\!\left(
\mathrm{ReLU}\!\left(\mathrm{BN}\!\left(
\mathrm{PW}_{1\times1}\!\left(\mathrm{Concat}(\mathbf{b}_1,\mathbf{b}_2)
\right)\right)\right)\right),
\label{eq:stage1}
\end{align}

\subsubsection{Downsample}

The downsample stage reduces the spatial resolution of the encoder
representation to a compact bottleneck through two successive strided
DW-Sep blocks. Enc-2 halves the spatial dimensions from $32\times32$
to $16\times16$ and its output is retained as skip connection that is denoted by
$\mathbf{s}_2 \in \mathbb{R}^{16\times16\times32}$. Enc-3 further
reduces the resolution to $8\times8$, producing the bottleneck
representation $\mathbf{x}_\text{bn}$. The downsample stage is
expressed using \ref{eq:stage2}.The $\mathbf{x}_\text{bn} \in \mathbb{R}^{8\times8\times32}$ is
the bottleneck feature map passed to the decoder, and the subscript
$s=2$ denotes strided convolution for spatial downsampling.

\begin{equation}
\label{eq:stage2}
    \mathbf{s}_2 = \mathrm{DWSep}_{s=2}(\mathbf{s}_1), \qquad
    \mathbf{x}_\text{bn} = \mathrm{DWSep}_{s=2}(\mathbf{s}_2)
\end{equation}

\subsubsection{Decoder}

The decoder restores spatial resolution through three successive bilinear
upsampling steps, following the Feature Pyramid Network
paradigm \cite{fpn}. At each stage, the upsampled feature map is
concatenated with the corresponding encoder skip connection and fused
by a $1\times1$ pointwise convolution before a DW-Sep block, recovering
spatial detail critical for accurate crop-row boundary delineation.
The decoder stage is expressed in \ref{eq:stage3}.

\begin{align}
\mathbf{d}_k &= \mathrm{DWSep}\!\left(
\mathrm{PW}_{1\times1}\!\left(
\mathrm{Concat}\!\left(
\mathrm{Up}(\mathbf{d}_{k+1}),\; \mathbf{s}_k
\right)\right)\right),
\label{eq:stage3}\\[-0.2em]
&\quad k \in \{2,1\} \nonumber
\end{align}

where $\mathbf{d}_{k+1}$ is the feature map from the previous decoder
stage (with $\mathbf{d}_3 = \mathbf{x}_\text{bn}$), $\mathbf{s}_k$ is
the corresponding skip connection from the encoder or downsample stage,
and $\mathrm{Up}(\cdot)$ denotes bilinear upsampling by a factor of 2.
The final decoder output $\mathbf{d}_0 \in \mathbb{R}^{64\times64\times16}$
is produced by a third upsampling step without a skip connection.

\subsubsection{Output}

A $1\times1$ convolution maps the final $64\times64\times16$ feature map
to $C = 2$ class logits, and a softmax activation converts these logits
into per-pixel class probabilities for the background and crop-row classes using \ref{eq:stage4}.

\begin{equation}
\label{eq:stage4}
    \hat{P}(y = c \mid \mathbf{x}) =
    \frac{\exp(z_c)}{\displaystyle\sum_{c'=0}^{C-1} \exp(z_{c'})},
    \quad c \in \{0, 1\}
\end{equation}

where $z_c$ denotes the logit for class $c$ produced by the $1\times1$
convolution, $C = 2$ is the total number of classes (background and
crop-row), and $\hat{P}(y = c \mid \mathbf{x})$ is the predicted probability
that pixel $\mathbf{x}$ belongs to class $c$.

\subsection{Training Objective}

The output probabilities are compared with the ground-truth
mask using a combined focal and Dice loss to address class
imbalance in crop-row segmentation. Focal loss emphasizes
hard foreground pixels, while Dice loss encourages overlap
between the predicted and ground-truth masks. The total loss
is defined in Eq.~\ref{eq:training_loss}.

\begin{equation}
    \mathcal{L} = \mathcal{L}_{\mathrm{focal}} + \mathcal{L}_{\mathrm{dice}}.
    \label{eq:training_loss}
\end{equation}

The model is trained using the Adam optimizer with an initial learning rate of $10^{-3}$ and a cosine decay learning-rate schedule. Checkpoints are selected based on the validation crop-row IoU. Multiple training durations are evaluated to analyze how segmentation quality changes with the number of epochs and the results are presented in Section \ref{sec:epoch}. Once training is complete, the optimised model weights are applied during inference. To further optimize this process, the network incorporates the two-phase progressive resolution training strategy established in paper \cite{sha1}, which initiates training with low-resolution inputs (64x64) before progressing to higher resolutions. This progressive training approach enables highly efficient parameter utilization, accelerates convergence, and reduces total training time by 28\%.

\subsection{Inference and Post-Processing}

At inference time, the RGB image is resized to $64 \times 64$ and passed through UltraLight Luma. The network produces a $64 \times 64 \times 2$ probability tensor, and the crop-row probability channel is bilinearly upsampled to the original image resolution. A fixed threshold $\tau = 0.5$ is then applied to obtain the binary
crop-row mask, as given in \ref{eq:binary_mask}.

\begin{equation}
    M_{\mathrm{bin}}(u,v) =
    \begin{cases}
        1, & \hat{P}_{\mathrm{crop}}(u,v) \geq 0.5, \\
        0, & \text{otherwise}
    \end{cases}
    \label{eq:binary_mask}
\end{equation}

where $(u,v)$ denotes the pixel coordinates in the upsampled map and $\hat{P}_{\mathrm{crop}}(u,v)$ is the upsampled crop-row probability at that position. For row-geometry analysis, the binary mask is skeletonized to obtain single-pixel-wide row centerlines. A small morphological dilation is then applied to make the extracted lines easier to visualize while preserving their geometric location. The resulting mask is used in the TSM-based analysis described in Section~\ref{sec:tsm}.




\section{EXPERIMENTAL EVALUATION}
\label{sec:rd}

This section evaluates the proposed model using quantitative segmentation metrics, qualitative prediction results, TSM-based crop-row analysis, and comparisons with YOLO and U-Net where appropriate. Experiments were conducted on a Ryzen 7 CPU and an NVIDIA RTX 4050 GPU with 4 GB of memory using the dataset described in Section~\ref{sec:ds}. Segmentation performance is measured using IoU across training epochs, while qualitative results assess robustness under selected field conditions.

\subsection{Training Progress and Segmentation Accuracy}
\label{sec:epoch}
\begin{figure}
\centering
\begin{tabular}{ccc}
  \includegraphics[width=0.25\linewidth]{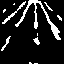} &
  \includegraphics[width=0.25\linewidth]{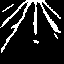} &
  \includegraphics[width=0.25\linewidth]{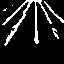} \\
  (a) & (b) & (c) \\[4mm] 

  \includegraphics[width=0.25\linewidth]{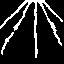} &
  \includegraphics[width=0.25\linewidth]{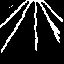} &
  \includegraphics[width=0.25\linewidth]{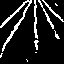} \\
  (d) & (e) & (f) \\ 
\end{tabular}

\caption{Crop-row segmentation predictions produced by UltraLight Luma at different training stages: (a) 50 epochs, (b) 100 epochs, (c) 150 epochs, (d) 200 epochs, (e) 250 epochs, and (f) 300 epochs. White pixels indicate predicted crop-row regions.}
\label{fig:epoch}
\end{figure}

The proposed model was trained for 50, 100, and 150 epochs to analyze the effect of training duration on crop-row segmentation performance. The prediction results at each training stage are presented in Fig.~\ref{fig:epoch}, where the detected crop-row regions are shown as white pixels in the binary output images. To quantitatively evaluate detection performance on the test dataset, IoU was used as the evaluation metric. The model achieved IoU values of 15.63\%, 17.77\%, 18.72\%, 19.66\%, 19.88\%, and 20.68\% after 50, 100, 150, 200, 250, and 300 epochs, respectively. These results show that segmentation accuracy improved as training progressed. The total training time for 300 epochs was 26.08 minutes, with a mean training time of 5.22 seconds per epoch. This average was computed across six training runs corresponding to 50, 100, 150, 200, 250, and 300 epochs.

\subsection{Crop Row Prediction Under Diverse Field Conditions}

\begin{figure}
\centering
\subfloat[RGB image]{\includegraphics[width=0.24\textwidth]{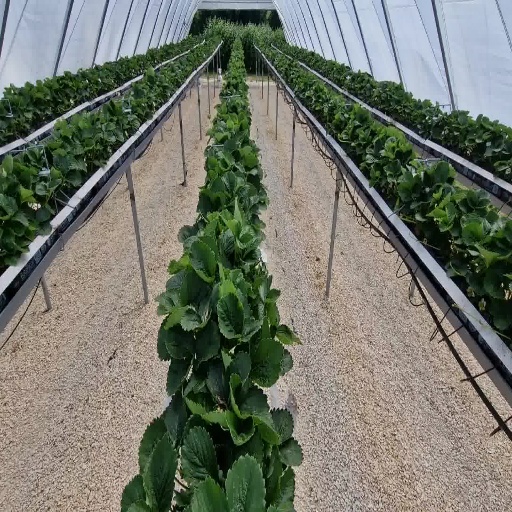}}\hfill
\subfloat[U-Net prediction]{\includegraphics[width=0.24\textwidth]{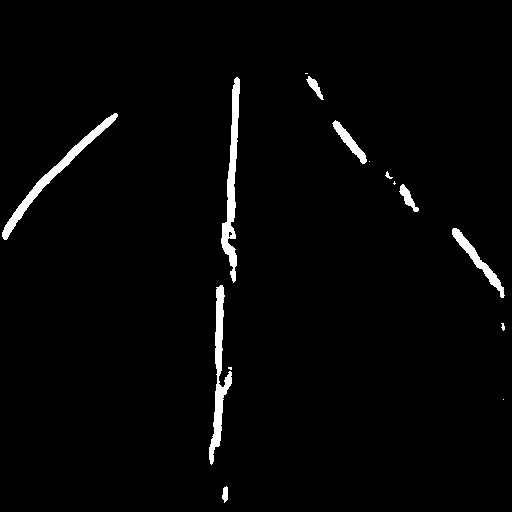}}\hfill
\subfloat[YOLO prediction]{\includegraphics[width=0.24\textwidth]{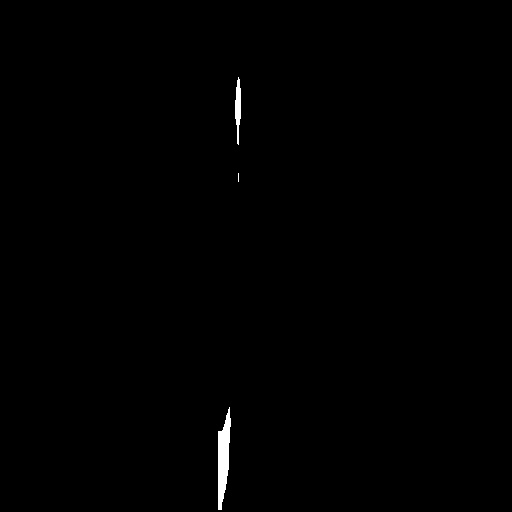}}\hfill
\subfloat[UltraLight Luma prediction]{\includegraphics[width=0.24\textwidth]{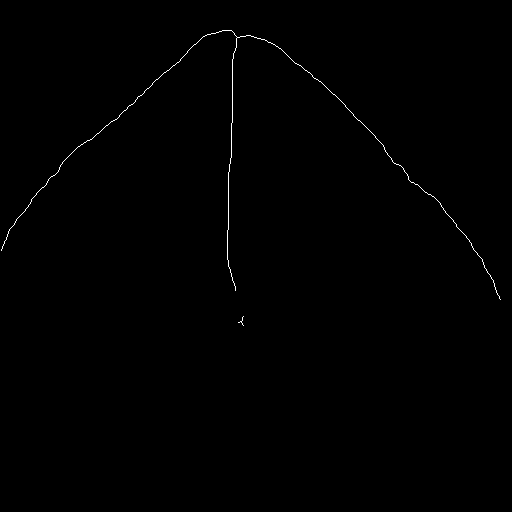}}
  \caption{Qualitative comparison of crop-row detection for a representative strawberry greenhouse image: (a) input RGB image, (b) U-Net segmentation prediction, (c) YOLOv8-based prediction, and (d) UltraLight Luma prediction.}
  \label{fig:qualitative_comparison}
\end{figure}

Figure~\ref{fig:tsm_output} shows the crop-row masks predicted by the Luma model for land crops such as maize and sugar beet, as well as greenhouse crops such as strawberry. Prediction under shadowed field conditions is also shown in Fig. ~\ref{fig:tsm_output}(d), indicating the model's ability to maintain crop-row detection under changes in illumination and scene appearance. Figure~\ref{fig:qualitative_comparison} compares crop-row predictions produced by U-Net~\cite{l1}, YOLOv8, and Luma from the same RGB input image. The qualitative results show that the proposed Luma model is capable of segmenting crop-rows efficiently. In addition, effective segmentation performance was observed under indoor greenhouse conditions.

\subsection{TSM-Based Crop Row Analysis}
\label{sec:tsm}

\begin{figure}
\centering
\subfloat[Maize]{\includegraphics[width=0.48\linewidth]{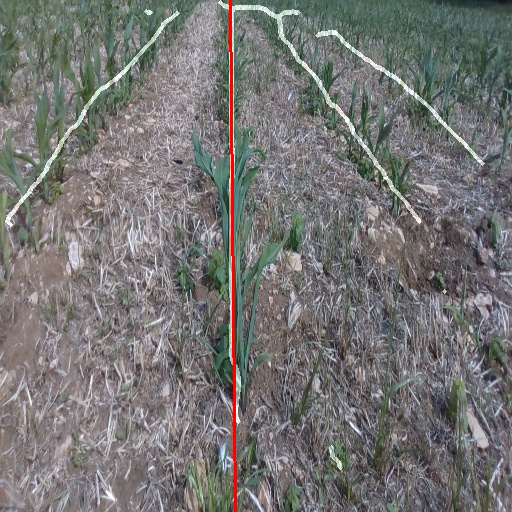}}%
\subfloat[Strawberry greenhouse]{\includegraphics[width=0.48\linewidth]{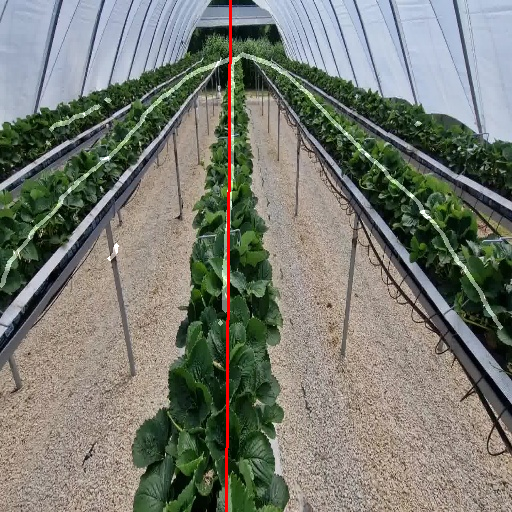}}\\
\subfloat[Sugar beet]{\includegraphics[width=0.48\linewidth]{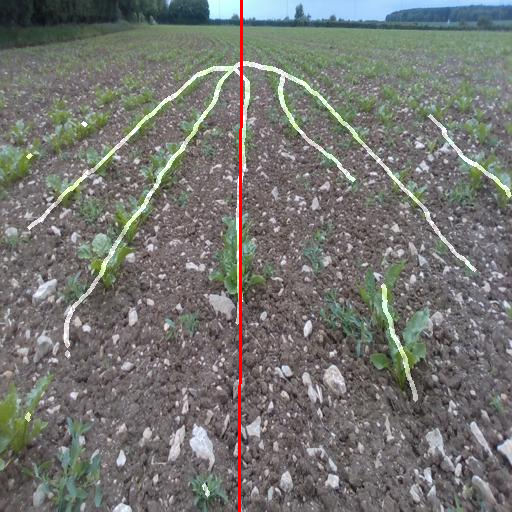}}%
\subfloat[Sugar beet with front shadows]{\includegraphics[width=0.48\linewidth]{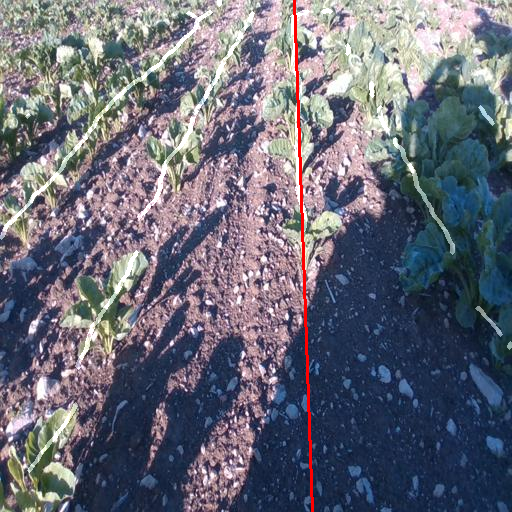}}
  \caption{Representative TSM-based crop-row detection outputs under different field conditions: (a) maize, (b) strawberry greenhouse, (c) sugar beet, and (d) sugar beet with front shadows. Each subplot shows the crop-row centerline estimated from the segmentation output and used to compute the angular deviation and lateral offset metrics.}
  \label{fig:tsm_output}
\end{figure}

The TSM is a post-processing algorithm that exploits the crop-row vanishing point caused by perspective distortion to define the top point of the centerline. Keeping this top point fixed, it performs a sweeping scan to candidate points along the bottom edge, and the line with the most white pixels is selected as the crop-row centerline, defining its angle and lateral position. The TSM was applied to evaluate the crop-row detection performance of the proposed model. For each image, two parameters were extracted from the detected crop-row: the angular deviation from the vertical axis ($\Delta\theta$) and the lateral offset of the lower endpoint from the image center ($\Delta L_{x2}$). These parameters were used to compute the overall detection performance $\epsilon$ as defined in~\ref{eq:epsilon}.

\begin{equation}
    \epsilon = 1 - \sum_{i=0}^{N} \frac{1}{2N}
    \left( \frac{\Delta\theta_i}{\Delta\theta_{max}} +
    \frac{\Delta L_{x2,i}}{\Delta L_{x2,max}} \right)
    \label{eq:epsilon}
\end{equation}

\noindent where $N$ is the total number of images evaluated,
$\Delta\theta_{max}$ and $\Delta L_{x2,max}$ are the maximum
observed angular deviation and lateral offset respectively,
used to normalize the errors within a consistent scale. Figure~\ref{fig:tsm_output} illustrates representative TSM
detection outputs across different field conditions. These outputs show that the Luma model can identify the navigation line from its predictions using the TSM algorithm. The values of $\Delta\theta_{max}$ and $\Delta L_{x2,max}$ were selected from the baseline defined in \cite{DESILVA2024108581}. The proposed Luma model achieved an average $\epsilon$ score of 74.81\%. Although it achieved a lower raw $\epsilon$ score than U-Net, it required substantially fewer trainable parameters, demonstrating stronger parameter efficiency. Therefore, a comparative analysis was conducted between the proposed Luma model and U-Net \cite{DESILVA2024108581}, YOLOv8, and YOLOv26 using the log-normalized parameter efficiency metric defined in Eq. \ref{eq:logN}. This metric enables a fairer comparison across models with considerably different parameter counts.

\begin{figure}
  \centering
  \includegraphics[width=0.98\columnwidth]{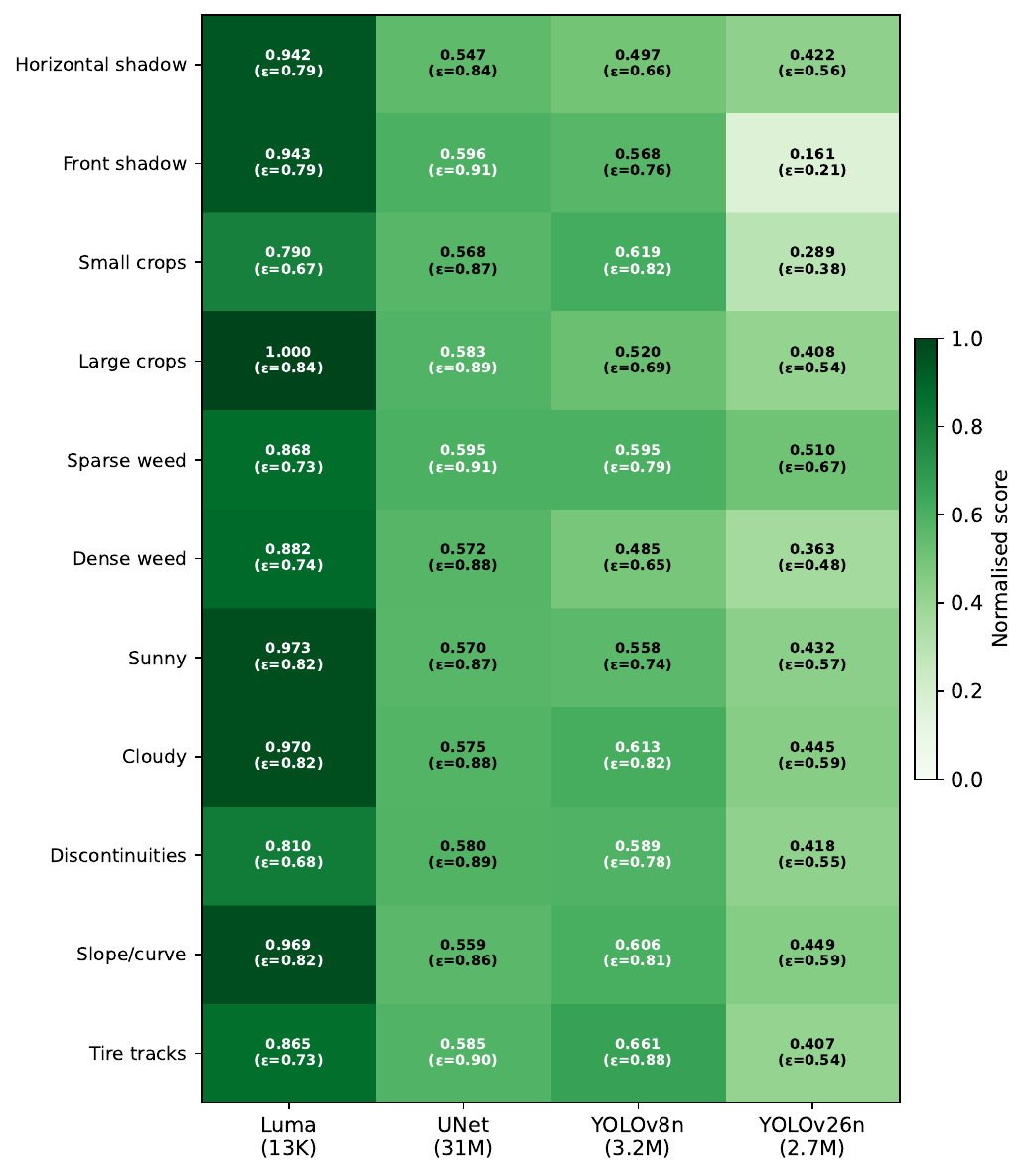}
  \caption{Log-parameter-normalized performance comparison of the proposed Luma model, U-Net, YOLOv8n, and YOLOv26n across different data categories. Each cell represents the normalized log-parameter-adjusted score, while the raw $\varepsilon$ value is shown in brackets. Higher values indicate better parameter-efficient performance.}
 \label{fig:heatmapTSM}
\end{figure}


\begin{equation}
\eta_{i,j} = \frac{\varepsilon_{i,j} \;/\; \log_{10}(P_i)}
                  {\max_{i,j}\left[\varepsilon_{i,j} \;/\; \log_{10}(P_i)\right]}
\label{eq:logN}
\end{equation}

\noindent where $\eta_{i,j} \in [0,1]$ is the log-normalized parameter efficiency
score, $\varepsilon_{i,j}$ is the raw performance score, and $P_i$ is the parameter
count of method $i$ in millions. The index $i \in \{\text{Luma, UNet, YOLOv8n,
YOLOv26n}\}$ denotes the segmentation method, and $j$ denotes the data category
as defined in \cite{DESILVA2024108581} and illustrated in Fig.~\ref{fig:heatmapTSM}, where each category is
represented along the x-axis. The constant $\delta = 10^{-6}$ ensures numerical stability, and the $\min$ and $\max$ operators are computed globally over all $i$ and $j$. The comparative analysis of $\eta_{i,j}$ values is presented in Fig.~\ref{fig:heatmapTSM}. The average $\eta_{i,j}$ values were 57.6\%, 57.4\%, and 39.1\% for U-Net, YOLOv8, and YOLOv26, respectively, whereas Luma achieved 91.0\%, demonstrating its superior performance. These results indicate
that the detected crop-rows are consistently well-centered
and aligned with the vertical axis, demonstrating the
capability of the proposed model to reliably detect crop
rows across varying field conditions.


\subsection{Computational Efficiency Analysis}

\begin{figure}
\centering
\subfloat[]{\includegraphics[width=0.48\linewidth]{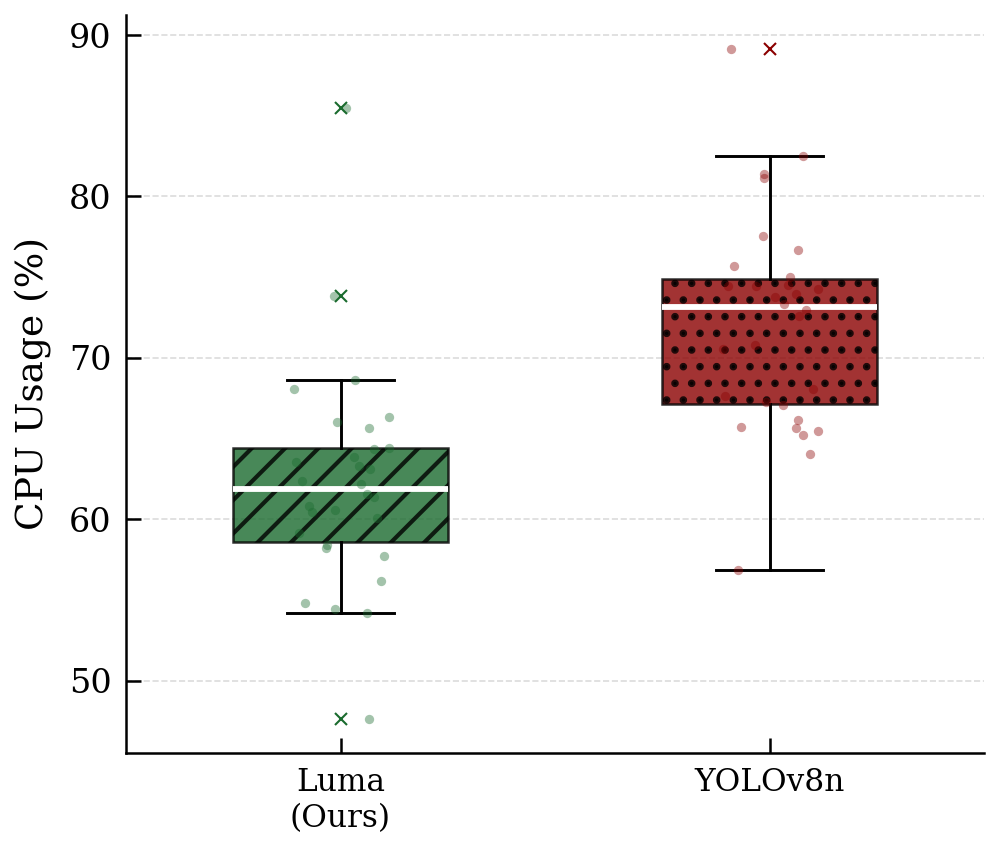}}%
\subfloat[]{\includegraphics[width=0.48\linewidth]{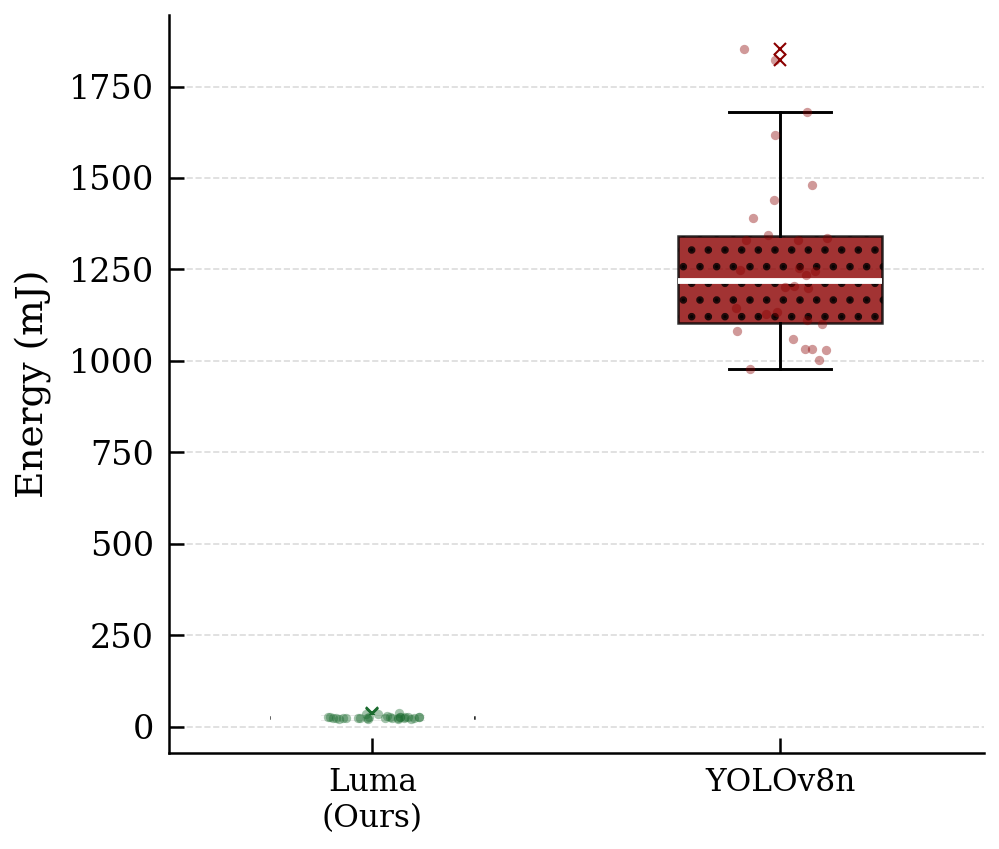}}\\
\subfloat[]{\includegraphics[width=0.48\linewidth]{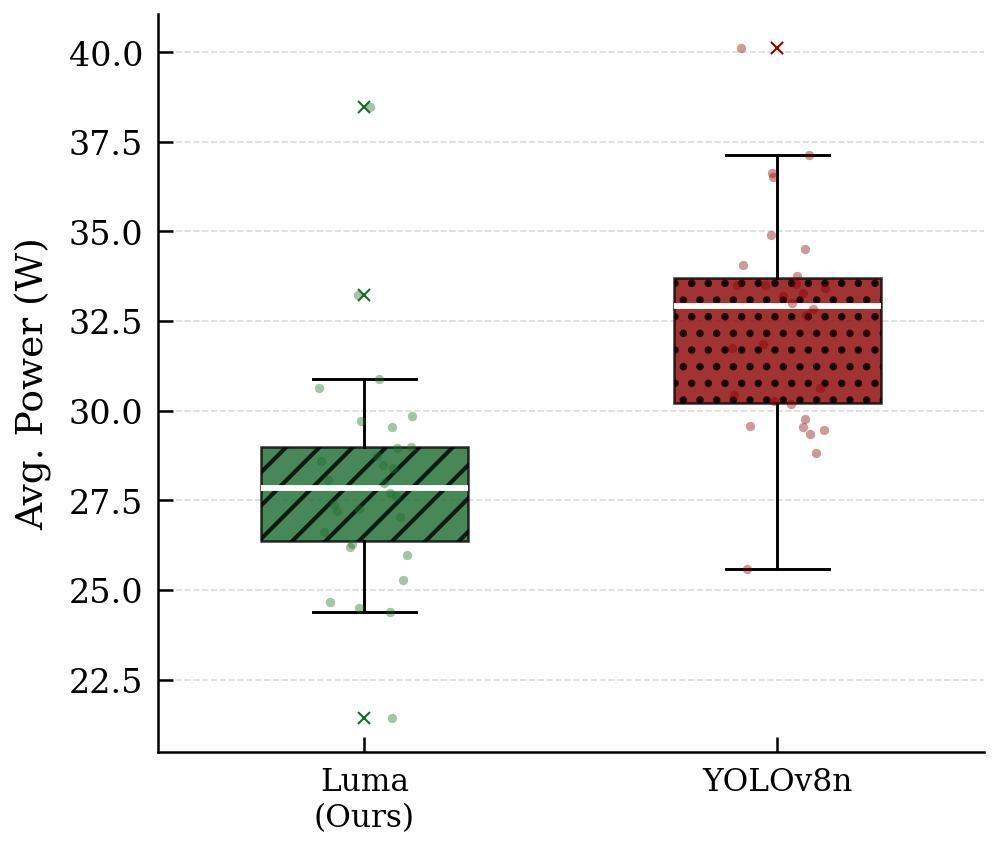}}%
\subfloat[]{\includegraphics[width=0.48\linewidth]{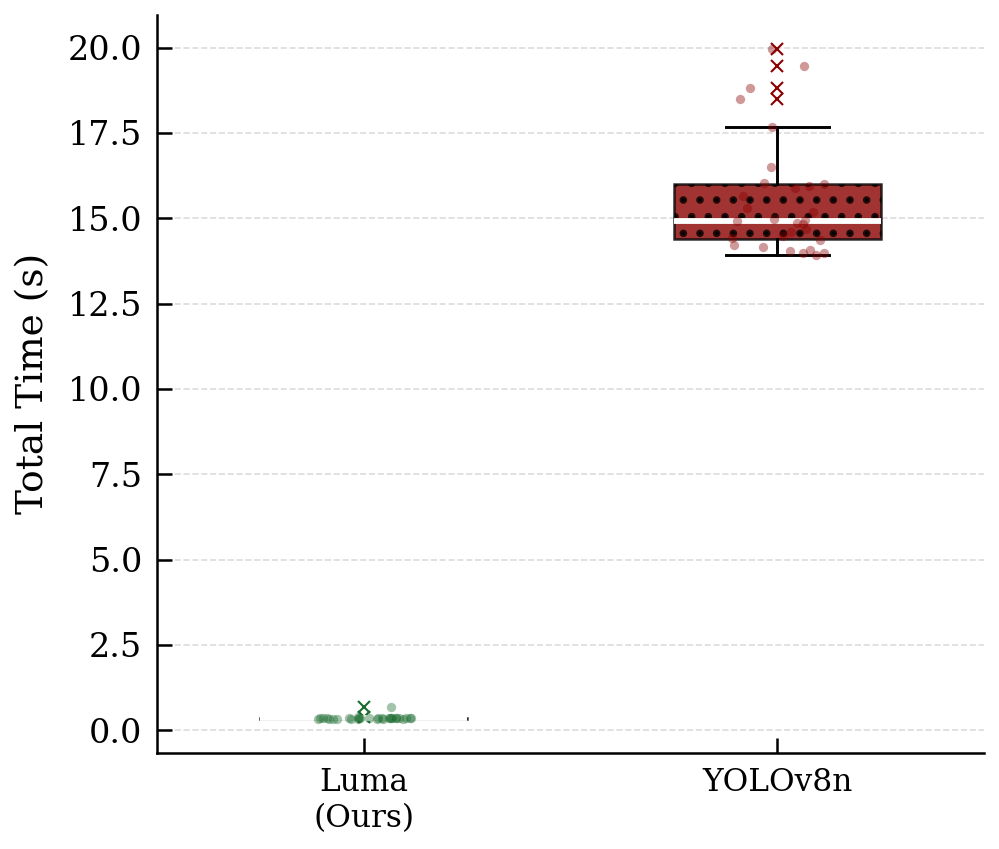}}

\begin{tabular}{cc}

\begin{tikzpicture}[baseline=-0.5ex]
  \draw[green!40!black, line width=1.5pt] (0,0) -- (0.6,0);
  \node[green!40!black] at (0.3,0) {\small$\blacksquare$};
\end{tikzpicture}
&
\begin{tikzpicture}[baseline=-0.5ex]
  \draw[red!70!black, line width=1.5pt] (0,0) -- (0.6,0);
  \node[red!70!black] at (0.3,0) {\small$\blacksquare$};
\end{tikzpicture}

\\

UltraLight Luma & YOLOv8n

\end{tabular}

\caption{Comparison of inference efficiency metrics between UltraLight Luma and YOLOv8n
over 30 inference cycles. UltraLight Luma consistently outperforms YOLOv8n across all
metrics, demonstrating significantly lower resource consumption.
(a) CPU utilization (\%), (b) energy per inference (mJ),
(c) average power consumption (W), and (d) total inference time (s).}
\label{fig:inference_comparison}
\end{figure}
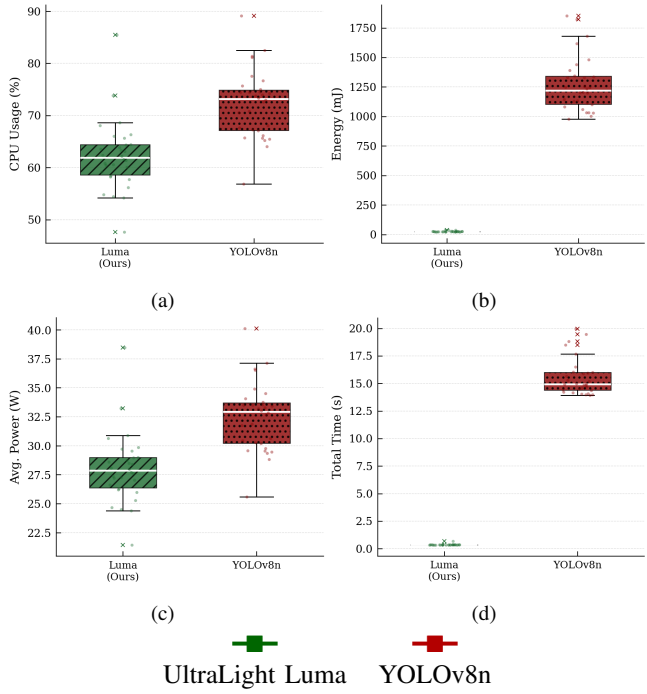

The inference efficiency of UltraLight Luma was compared against YOLOv8n across
four metrics: CPU utilization, energy per inference, average power consumption,
and total inference time, measured over 30 test cases. As shown in
Fig.~\ref{fig:inference_comparison} and Table~\ref{tab:inference_comparison},
UltraLight Luma achieved lower resource consumption across all metrics, with only
0.013M parameters compared to 3.157M for YOLOv8n, making it approximately 243
times smaller while requiring less energy and power during inference.

\begin{table}[h]
\centering
\caption{Inference efficiency comparison: UltraLight Luma vs. YOLOv8n.}
\label{tab:inference_comparison}
\renewcommand{\arraystretch}{1.25}
\setlength{\tabcolsep}{2.2pt}
\scriptsize
\resizebox{\columnwidth}{!}{%
\begin{tabular}{|l|c|c|c|c|c|}
\hline
\textbf{Model} & \textbf{Params} & \textbf{Size} & \textbf{CPU} & \textbf{Power} & \textbf{Energy} \\
 & \textbf{(M)} & \textbf{(MB)} & \textbf{(\%)} & \textbf{(W)} & \textbf{(mJ/inf.)} \\
\hline
UltraLight Luma & 0.013 & 0.517 & 62.23 & 28.00 & 25.23 \\
\hline
YOLOv8n & 3.157 & 6.550 & 72.13 & 32.46 & 1268.50 \\
\hline
\end{tabular}%
}
\end{table}


\section{Conclusion}
\label{sec:conclusion}

This paper presented UltraLight Luma, an ultra-lightweight encoder-decoder segmentation network for crop-row perception on low-cost agricultural robotic platforms. The proposed approach formulates crop-row detection as a binary segmentation problem and combines the predicted mask with a Triangle Scan Method-based navigation-line extraction module to support downstream visual servoing. Despite its modest maximum IoU of 20.68\%, UltraLight Luma preserved sufficient crop-row structure to achieve an average crop-row detection score of 74.81\%, while using only 13.1k trainable parameters and reaching a parameter-efficiency score of 91.0\%. These results show that UltraLight Luma offers a practical balance between segmentation performance, navigation-line extraction, and edge-deployable efficiency for resource-constrained agricultural robots.

\section{Acknowledgement}
\label{sec:acknowledgement}

The authors acknowledge the use of OpenAI ChatGPT for language editing, grammar improvement, and clarity enhancement during manuscript preparation. The AI tool was used only to refine the presentation of author-provided technical content and final manuscript content were reviewed and verified by the authors.


\bibliographystyle{ieeetr}

\bibliography{glorified,new}

@STRING{arxiv   = {arXiv preprint} }

@STRING{sensors = {Sensors} }

@STRING{springer= {Springer Verlag} }

@INPROCEEDINGS{10593228,
  author={Balasingham, Dhanushka and Samarathunga, Sadeesha and Arachchige, Gayantha Godakanda and Bandara, Anuththara and Wellalage, Sasini and Pandithage, Dinithi and Hansika, Mahaadikara M.D.J.T and De Silva, Rajitha},
  booktitle={2024 5th International Conference for Emerging Technology (INCET)}, 
  title={SPARROW: Smart Precision Agriculture Robot for Ridding of Weeds}, 
  year={2024},
  volume={},
  number={},
  pages={1-6},
  doi={10.1109/INCET61516.2024.10593228}}

@INPROCEEDINGS{9644466,
  author={Hu, Yue and Huang, He},
  booktitle={2021 6th International Symposium on Computer and Information Processing Technology (ISCIPT)}, 
  title={Extraction Method for Centerlines of Crop Row Based on Improved Lightweight Yolov4}, 
  year={2021},
  volume={},
  number={},
  pages={127-132},
  doi={10.1109/ISCIPT53667.2021.00032}}

@INPROCEEDINGS{10021091,
  author={Doha, Rashed and Hasan, Mohammad Al and Anwar, Sohel},
  booktitle={2022 IEEE International Conference on Big Data (Big Data)}, 
  title={Semantic Segmentation Approaches in Crop Row Detection}, 
  year={2022},
  volume={},
  number={},
  pages={2224-2230},
  doi={10.1109/BigData55660.2022.10021091}}

@article{l1,
author = {de Silva, Rajitha and Cielniak, Grzegorz and Wang, Gang and Gao, Junfeng},
title = {Deep learning-based crop row detection for infield navigation of agri-robots},
journal = {Journal of Field Robotics},
volume = {41},
number = {7},
pages = {2299-2321},
doi = {https://doi.org/10.1002/rob.22238},
url = {https://onlinelibrary.wiley.com/doi/abs/10.1002/rob.22238},
eprint = {https://onlinelibrary.wiley.com/doi/pdf/10.1002/rob.22238},
year = {2024}
}

@article{LI2024109264,
title = {Rethinking the crop row detection pipeline: An end-to-end method for crop row detection based on row-column attention},
journal = {Computers and Electronics in Agriculture},
volume = {225},
pages = {109264},
year = {2024},
issn = {0168-1699},
doi = {https://doi.org/10.1016/j.compag.2024.109264},
url = {https://www.sciencedirect.com/science/article/pii/S0168169924006550},
author = {Boliao Li and Dongfang Li and Zhenbo Wei and Jun Wang}
}

@article{linknet,
  author       = {Abhishek Chaurasia and
                  Eugenio Culurciello},
  title        = {LinkNet: Exploiting Encoder Representations for Efficient Semantic
                  Segmentation},
  journal      = {CoRR},
  volume       = {abs/1707.03718},
  year         = {2017},
  url          = {http://arxiv.org/abs/1707.03718},
  eprinttype   = {arXiv},
  eprint       = {1707.03718},
  bibsource    = {dblp computer science bibliography, https://dblp.org}
}

@article{enet,
title = {Study of convolutional neural network-based semantic segmentation methods on edge intelligence devices for field agricultural robot navigation line extraction},
journal = {Computers and Electronics in Agriculture},
volume = {209},
pages = {107811},
year = {2023},
issn = {0168-1699},
doi = {https://doi.org/10.1016/j.compag.2023.107811},
url = {https://www.sciencedirect.com/science/article/pii/S0168169923001990},
author = {Jiya Yu and Jiye Zhang and Aijing Shu and Yujie Chen and Jianneng Chen and Yongjie Yang and Wei Tang and Yanchao Zhang}
}

@ARTICLE{boundingBoxYolo,
    
AUTHOR={Gong, Hailiang  and Zhuang, Weidong  and Wang, Xi },
           
TITLE={Improving the maize crop row navigation line recognition method of YOLOX},
          
JOURNAL={Frontiers in Plant Science},
          
VOLUME={Volume 15 - 2024},
  
YEAR={2024},
  
URL={https://www.frontiersin.org/journals/plant-science/articles/10.3389/fpls.2024.1338228},
  
DOI={10.3389/fpls.2024.1338228},
  
ISSN={1664-462X}}

@Article{agriculture15070796,
AUTHOR = {Sun, Tao and Le, Feixiang and Cai, Chen and Jin, Yongkui and Xue, Xinyu and Cui, Longfei},
TITLE = {Soybean–Corn Seedling Crop Row Detection for Agricultural Autonomous Navigation Based on GD-YOLOv10n-Seg},
JOURNAL = {Agriculture},
VOLUME = {15},
YEAR = {2025},
NUMBER = {7},
ARTICLE-NUMBER = {796},
URL = {https://www.mdpi.com/2077-0472/15/7/796},
ISSN = {2077-0472},
DOI = {10.3390/agriculture15070796}
}

@Article{s26103142,
AUTHOR = {Wu, Letian and Cui, Yongzhi and Shi, Huifeng and Sun, Xiaoli and Yang, Jiayan and Cao, Xinwei and Zou, Ping and Liu, Ya},
TITLE = {An Improved DeepLabV3+-Based Method for Crop Row Segmentation and Navigation Line Extraction in Agricultural Fields},
JOURNAL = {Sensors},
VOLUME = {26},
YEAR = {2026},
NUMBER = {10},
ARTICLE-NUMBER = {3142},
URL = {https://www.mdpi.com/1424-8220/26/10/3142},
ISSN = {1424-8220},
DOI = {10.3390/s26103142}
}

@article{sze2017efficient,
  author = {Sze, Vivienne and Chen, Yu-Hsin and Yang, Tien-Ju and Emer, Joel S.},
  title = {Efficient Processing of Deep Neural Networks: A Tutorial and Survey},
  journal = {Proceedings of the IEEE},
  volume = {105},
  number = {12},
  pages = {2295--2329},
  year = {2017},
  doi = {10.1109/JPROC.2017.2761740}
}

@article{fpn,
  author       = {Tsung{-}Yi Lin and
                  Piotr Doll{\'{a}}r and
                  Ross B. Girshick and
                  Kaiming He and
                  Bharath Hariharan and
                  Serge J. Belongie},
  title        = {Feature Pyramid Networks for Object Detection},
  journal      = {CoRR},
  volume       = {abs/1612.03144},
  year         = {2016},
  url          = {http://arxiv.org/abs/1612.03144},
  eprinttype   = {arXiv},
  eprint       = {1612.03144},
  bibsource    = {dblp computer science bibliography, https://dblp.org}
}

@INPROCEEDINGS{ginigaddarage2025,
  author={Ginigaddarage, Shamod and Rajahrajasingh, Hanojhan},
  booktitle={2025 IEEE 19th International Conference on Industrial and Information Systems (ICIIS)}, 
  title={UltraLight+: An Optimized Progressive CNN Architecture for Efficient Image Classification}, 
  year={2026},
  volume={19},
  number={},
  pages={401-406},
  doi={10.1109/ICIIS69028.2026.11450640}}

@InProceedings{unet,
author="Ronneberger, Olaf
and Fischer, Philipp
and Brox, Thomas",
editor="Navab, Nassir
and Hornegger, Joachim
and Wells, William M.
and Frangi, Alejandro F.",
title="U-Net: Convolutional Networks for Biomedical Image Segmentation",
booktitle="Medical Image Computing and Computer-Assisted Intervention -- MICCAI 2015",
year="2015",
publisher="Springer International Publishing",
address="Cham",
pages="234--241",
isbn="978-3-319-24574-4"
}

@article{DESILVA2024108581,
title = {Vision based crop row navigation under varying field conditions in arable fields},
journal = {Computers and Electronics in Agriculture},
volume = {217},
pages = {108581},
year = {2024},
issn = {0168-1699},
doi = {https://doi.org/10.1016/j.compag.2023.108581},
url = {https://www.sciencedirect.com/science/article/pii/S0168169923009699},
author = {Rajitha {de Silva} and Grzegorz Cielniak and Junfeng Gao}
}

@inproceedings{sha1,
author = {Ginigaddarage, Shamod and Rajahrajasingh, Hanojhan},
year = {2026},
month = {01},
pages = {},
title = {UltraLight: Progressive Resolution Training for Ultra-Efficient CNNs on Edge Devices}
}

\end{document}